\documentclass[runningheads,orivec]{llncs}
\usepackage[T1]{fontenc}
\usepackage{graphicx}
\usepackage[nolist]{acronym}
\usepackage{amsmath,amssymb}
\usepackage{multirow}
\usepackage[locale=DE]{siunitx}

\usepackage{booktabs}

\begin{document}
\title{Reinforcement Learning for Optimal Experiment Design in Parameter Identification of Mechatronic Systems}
\titlerunning{Reinforcement Learning for Parameter Identification}
%
\author{Julian Langschwert\inst{1,2} \and
Georg Schäfer\inst{1,2} \and
Jakob Rehrl\inst{1} \and
Stefan Huber\inst{1} \and
Simon Hirländer\inst{2}}
\authorrunning{J. Langschwert et al.}

\institute{Josef Ressel Centre for Intelligent and Secure Industrial Automation, \\
Salzburg University of Applied Sciences, Salzburg, Austria \and
Paris Lodron University of Salzburg, Salzburg, Austria \\
\email{julian.langschwert@fh-salzburg.ac.at}}
\maketitle              
\begin{abstract}
Informative excitation signals are critical for accurate system identification of mechatronic systems, yet classical \ac{si} approaches require expert knowledge and hand-crafted signal design to respect hardware safety constraints, limiting their generalizability. We propose a \ac{rl} agent that learns optimal excitation signals for a Quanser Aero~2 testbed while autonomously enforcing safety constraints through reward shaping. Evaluated across 10 independent training seeds, our comprehensive agent achieves competitive estimation accuracy across all three identified parameters, outperforming classical baselines while incurring only $0.75\%$ safety violations.

\end{abstract}
\acresetall
\section{Introduction}
\Ac{si} is a fundamental process for modeling complex dynamics~\cite{ljung_system_1999}. Field experts with extensive system and domain knowledge are required for tasks such as model selection, \ac{oed}, data collection, and parameter estimation. A complicating factor is that many \acp{cps} have strict hardware or safety constraints that need to be respected when designing these experiments.
Current \ac{si} strategies rely on standard open-loop excitation signals such as \ac{prbs} or frequency sweeps~\cite{ljung_system_1999}. While highly effective, their application necessitates extensive manual tuning to simultaneously maximize information yield~\cite{mazhar_aircraft_2025} and respect strict physical safety constraints.
Recent works demonstrate that \ac{rl} can successfully automate \ac{oed}; for instance, by optimizing input sequences for parameter estimation in Lithium-ion batteries~\cite{huang_reinforcement_2023}, and by utilizing D-optimality-guided rewards to efficiently calibrate a multi-DOF rehabilitation robot~\cite{hu_d-optimality-guided_2026}. Building upon this concept, this work-in-progress paper contributes an \ac{rl} agent that learns an excitation signal for a Quanser Aero~2 testbed in a 1-\ac{dof} configuration while minimizing parameter estimation error and autonomously respecting hardware safety constraints through reward shaping. 

\section{Methodology and Experimental Setup}
\paragraph{System Description and Model Formulation.}
The Quanser Aero~2 is a dual-rotor testbed configured here for a 1-\ac{dof} pitch motion (yaw locked, rotors horizontal). It is driven by a single voltage input $u \in [-24, 24]$\,V, applied to the motors as $u_1 = u$ and $u_2 = -u$, resulting in a seesaw motion with a manufacturer-specified pitch limit of $\pm45^\circ$.

The dynamic behavior of the pitch axis is governed by a second-order nonlinear differential equation:
\begin{subequations}
    \begin{align}
                J_p \ddot{\Theta} &= (F_1 - F_2)l - D_p \dot{\Theta} - m g d_S \sin(\Theta) \\
        F_1 &= \frac{J_p}{l} K_{pu} u_1, \quad F_2 = \frac{J_p}{l} K_{pu} u_2
    \end{align}
\end{subequations}
where $\Theta$, $\dot{\Theta}$, and $\ddot{\Theta}$ represent the pitch angle, angular velocity, and angular acceleration, respectively. The terms $F_1$ and $F_2$ denote the thrust forces generated by the rotors, which are located at a distance $l$ from the pivot. While physical constants such as the system mass ($m$), gravity ($g$), and center of mass offset ($d_S$) are known, three core dynamic parameters remain unknown and are the target of our \ac{si}: the moment of inertia ($J_p$), the viscous damping coefficient ($D_p$), and the voltage-to-thrust gain ($K_{pu}$).

\paragraph{Optimal Experiment Design via Reinforcement Learning.}
We formulate the active excitation design as a finite-horizon \ac{mdp} defined by the tuple $(\mathcal{S}, \mathcal{A}, r, \gamma)$. \\
\indent \textit{State and Action Space:}
The agent observes a sliding window of the $N_w = 80$ most recent noisy angle measurements and applied voltages. It additionally receives a normalized time index $k/T$ to track progression across the $T = 500$ step episode, enabling the agent to adapt its excitation strategy over time:
\begin{equation}
    s_k = \left[ \frac{y_{k-N_w+1}}{y_{\max}}, \dots, \frac{y_k}{y_{\max}},\; \frac{u_{k-N_w+1}}{u_{\max}}, \dots, \frac{u_k}{u_{\max}},\; \frac{k}{T} \right]
\end{equation}



\noindent At each step $k$, the agent outputs a continuous action $a_k \in [-1, 1]$, scaled to the physical motor voltage limits of $\pm 24$\,V.
\indent \textit{Parameter Estimation:}
For parameter identification, a discrete \ac{arx} model coupled with a \ac{rls} estimator is utilized, recovering the physical parameters via a nonlinear inverse mapping at the end of the episode. \\
\indent \textit{Reward Formulation:}
The objective of the agent is to maximize the expected discounted return. The reward function is designed to balance parameter estimation quality with strict hardware safety, quantified by: \\
\indent \textit{1. Stepwise Safety Penalty:} The Quanser Aero~2 has a manufacturer-specified operating range of $\pm 45^\circ$. To account for system inertia and avoid mechanical stress at the physical boundaries, we define a conservative critical limit $\theta_{\lim} = 40^\circ$. Furthermore, a warning threshold of $\theta_{\mathrm{warn}} = 30^\circ$ was determined empirically to provide a sufficient $10^\circ$ buffer zone, allowing the agent to learn to decelerate before breaching the limit. The agent receives a quadratic penalty inside this warning zone and a severe penalty for breaching the critical limit:
\begin{equation}
    r_k^{\mathrm{safety}} = \begin{cases} -1 - 2\frac{|\theta_k| - \theta_{\lim}}{\theta_{\lim}} & \text{if } |\theta_k| \geq \theta_{\lim}, \\ -0.1 \left(\frac{|\theta_k| - \theta_{\mathrm{warn}}}{\theta_{\lim} - \theta_{\mathrm{warn}}}\right)^2 & \text{if } |\theta_k| \geq \theta_{\mathrm{warn}}, \\ 0 & \text{otherwise}. \end{cases}
\end{equation}
\indent \textit{2. Terminal Estimation Reward:} Crucially, the \ac{rl} agent is trained entirely in a simulation environment. During this training phase, the physical parameters are based on nominal estimates from prior baseline experiments. In simulation, these parameters serve as the known ground truth $p_i^*$ for the reward calculation, allowing the agent to learn an optimal policy offline that can subsequently be deployed on real hardware where the parameters are truly unknown. Because the estimation quality is evaluated at the end of the trajectory, the agent receives a sparse terminal reward based on the relative error between the \ac{rls}-estimated parameters $\hat{p}_{i,T}$ and the simulated ground truth $p_i^*$:
\begin{equation}
    r_T^{\mathrm{est}} = -\sum_{p_i \in \mathcal{P}_{\mathrm{target}}} \frac{|\hat{p}_{i,T} - p_i^*|}{p_i^*}
\end{equation}
\paragraph{Training and Evaluation Setup.}
The \ac{rl} agent uses a \ac{ppo} algorithm from Stable-Baselines3 with two hidden layers of 64 neurons each. To ensure robust learning and address the sim-to-real gap, the true physical parameters are randomized by $\pm30\%$ of their nominal values at the start of each episode. To thoroughly evaluate the proposed approach, four different \ac{rl} agents were trained. Three specialized agents were trained to prioritize the estimation accuracy of a single specific parameter ($J_p$, $K_{pu}$, or $D_p$). While the underlying \ac{rls} algorithm always estimates all three parameters, this modification was achieved by changing the terminal reward to penalize only the relative error of the targeted parameter. A fourth, comprehensive agent was trained to jointly estimate all three parameters using the full terminal reward formulation. Each agent was trained with ten independent random seeds to ensure robustness to training stochasticity. \\
\indent \textit{Baseline Comparison:}
To evaluate the \ac{rl} approach, the trained agents are benchmarked against three classical open-loop signals, all restricted to a conservative $\pm7.2$\,V amplitude to ensure hardware safety. The first is a naive \textbf{random} excitation drawn uniformly from $[-7.2\,\mathrm{V}, 7.2\,\mathrm{V}]$ at each step. The second is a persistently exciting \textbf{\ac{prbs}} switching between discrete $\pm7.2$\,V states. The third is a \textbf{Chirp} signal sweeping from $0.05$ to $2$\,Hz, providing broadband excitation tailored to the pitch dynamics. For a strictly fair comparison, all methods are evaluated at a $10$\,Hz sampling rate over the exact same episode length using the identical least-squares algorithm.

\section{Results and Discussion}
\textit{Training Stability.} As detailed in Table~\ref{tab:results_multiseed}, the $N_{\mathrm{seeds}}$ column indicates that targeting $D_p$ makes training less stable. While the parameter-specific agents \ac{rl}($J_p$) and \ac{rl}($K_{pu}$) achieved full convergence, \ac{rl}($D_p$) converged in 9/10 seeds, and \ac{rl}(all) in only 8/10. Specifically, two of the comprehensive agents collapsed to a zero-output policy, highlighting the challenge of jointly optimizing for all three parameters simultaneously. Furthermore, the observed \acp{sd} are generally large. For the $J_p$ estimation of the \ac{rl}($K_{pu}$) agent, the \ac{sd} entirely exceeds the mean, underlining the need for more stable training procedures in future work. \\
\indent \textit{Estimation Accuracy and Safety Trade-off.}
The \ac{rl}($D_p$) agent outperforms all other policies across \textit{all three} parameters, suggesting that successfully exciting the most difficult parameter yields a trajectory rich enough to identify the remaining ones. However, the \ac{rl}($D_p$) agent's regular safety violations of 151/900 evaluations (a rate of approximately 16.8\%) render it unsuitable for deployment on physical hardware. Consequently, the comprehensive agent emerges as the practically preferred policy, achieving $D_p$ error of 198\% versus 495\% for the best classical baseline while safely operating within physical limits, incurring only 0.75\% safety violations. \\
\indent \textit{Damping Identification Challenges.}
Despite these results, all methods continue to struggle with accurately identifying the viscous damping coefficient $D_p$. This is consistent with ill-conditioning in the discrete \ac{arx} regressor matrix: $D_p$ contributes weakly to the regression signal relative to $J_p$ and $K_{pu}$, making its contribution difficult to isolate regardless of excitation quality. \\
\indent \textit{Actuator Range and Comparison Validity.}
A key advantage of the \ac{rl} approach is its ability to dynamically utilize the full $\pm24$\,V range while respecting safety constraints through reward shaping. Classical baselines required manual restrictions to $\pm7.2$\,V to prevent safety violations. Yet, even with this conservative amplitude, the \ac{prbs} baseline still exceeded the physical limit twice. While this disparity 
in operating voltage limits direct comparison, it simultaneously demonstrates the core contribution of \ac{rl}-based \ac{oed}: rather than requiring conservative hand-crafted amplitude limits, the agent learns to safely exploit the available actuator range to maximize information yield. \\
\begin{table}[t]
  \centering
  \caption{Robustness across 10 training seeds ($N_{\mathrm{seeds}}$ denotes viable seeds), evaluated on 100 randomized episodes. Values report the cross-seed mean $\pm$ std of the per-seed mean absolute relative error (\acs{mare}) of parameters and mean absolute maximum angles during evaluation.}
  \label{tab:results_multiseed}
  \begin{tabular*}{\textwidth}{@{\extracolsep{\fill}}l r@{\,$\pm$\,}l r@{\,$\pm$\,}l r@{\,$\pm$\,}l c r@{\,$\pm$\,}l c@{}}
    \toprule
    \textbf{Policy} & 
    \multicolumn{2}{c}{\textbf{$J_p$ / \%}} & 
    \multicolumn{2}{c}{\textbf{$K_{pu}$ / \%}} & 
    \multicolumn{2}{c}{\textbf{$D_p$ / \%}} & 
    {\footnotesize \textbf{Safety Viol.}} & 
    \multicolumn{2}{c}{\textbf{$\theta_{\max}$ / $^\circ$}} &
    $N_{\mathrm{seeds}}$ \\
    \midrule
    \multicolumn{10}{@{}l}{\textbf{Baselines} \textit{($\pm 7.2$\,V)}} \\[2pt]
    Random    & 8.23 & 6.61 & 10.18 & 5.14 & 3026 & 2067  & 0/100 & 11.2 & 4.6 & --- \\
    PRBS      & 2.57 & 2.73 &  3.42 & 1.64 & 1065 & 936  & 2/100 & 20.4 & 9.1 & --- \\
    Chirp     & 1.43 & 0.57 &  2.65 & 1.09 &  495 & 202  & 0/100 & 24.4 & 4.3 & --- \\
    \midrule
    \multicolumn{10}{@{}l}{\textbf{RL Agents} \textit{($\pm 24$\,V, cross-seed statistics)}} \\[2pt]
    RL($J_p$)    & 1.87 & 1.38 & 1.86 & 1.15 & 4577 & 2436 & 6/1000 & 17.0 & 4.7 & 10/10 \\
    RL($K_{pu}$) & 5.70 & 5.79 & 0.40 & 0.36 & 1727 & 1179 & 3/1000 & 16.9 & 5.6 & 10/10 \\
    RL($D_p$)    & \textbf{0.75} & \textbf{0.38} & \textbf{0.37} & \textbf{0.20} & \textbf{190} & \textbf{86} & 151/900 & 36.1 & 6.0 & 9/10 \\
    RL(all)      & 1.54 & 1.04 & 0.51 & 0.40 & 198 & 136 & 6/800 & 26.5 & 2.0 & 8/10 \\
    \bottomrule
  \end{tabular*}
\end{table}
\indent \textit{Conclusion and Future Work.}
This work demonstrates that \ac{rl} can successfully automate optimal experimental design for mechatronic systems by learning excitation signals that outperform classical baselines while autonomously enforcing hardware safety constraints. However, accurately identifying $D_p$ remains an open challenge across all methods. To address this, we will replace the discrete ARX-RLS estimator with a batch \ac{nls} approach operating on the continuous-time trajectory, avoiding ill-conditioning of the inverse mapping. In parallel, we plan to replace the sparse terminal reward with a dense, step-wise reward formulated around the Fisher Information Matrix, which directly measures the information content of the excitation trajectory. Further simulation work will refine the safety formulation by normalizing the penalty with respect to episode length $T$, investigate broader parameter randomization bounds to improve policy robustness, and conduct a formal optimization of the classical baseline parameters for a more rigorous comparison. Once the simulation framework is mature, the learned policies will be validated on the physical Quanser Aero~2 hardware to evaluate sim-to-real transfer.

\paragraph{Acknowledgments.}
Financial support for this study was provided by the Christian Doppler Research Association (CDG) through the Josef Ressel Centre for Intelligent and Secure Industrial Automation, the corresponding WISS Co-project of Land Salzburg, and by the European Interreg project BA0100172 AI4GREEN.
%
%
%
\bibliographystyle{splncs04}
\bibliography{references}
\begin{acronym}
    \acrodef{rl}[RL]{reinforcement learning}
    \acro{pem}[PEM]{prediction error minimization}
    \acro{cps}[CPS]{cyber-physical system}
    \acro{mpc}[MPC]{model predictive control}
    \acro{dof}[DOF]{degree-of-freedom}
    \acro{rls}[RLS]{recursive least-squares}
    \acro{pem}[PEM]{prediction error method}
    \acro{oed}[OED]{optimal experimental design}
    \acro{si}[SI]{system identification}
    \acro{prbs}[PRBS]{pseudo-random binary sequences}
    \acro{mdp}[MDP]{Markov decision process}
    \acrodef{ppo}[PPO]{proximal policy optimization}
    \acrodef{mlp}[MLP]{multi layer perceptron}
    \acrodef{arx}[ARX]{AutoRegressive with eXogenous inputs}
    \acrodef{nls}[NLS]{Nonlinear Least Squares}
    \acrodef{mare}[MARE]{mean absolute relative error}
    \acrodef{sd}[SD]{standard deviation}
\end{acronym}

\end{document}